\documentclass[lettersize,journal]{IEEEtran}
\usepackage{times}
\usepackage{epsfig}
\usepackage{booktabs}
\usepackage{tabularx}
\usepackage{multirow}
\usepackage{subcaption}
\usepackage[export]{adjustbox}
\usepackage{float}
\usepackage{dashrule}
\usepackage{arydshln}
\usepackage[utf8x]{inputenc}
\usepackage[space]{grffile}
\usepackage{anyfontsize}
\usepackage{t1enc}
\usepackage{mathtools, nccmath}

\usepackage{amsmath,amsfonts}
\usepackage{algorithmic}
\usepackage{algorithm}
\usepackage{array}
\usepackage{textcomp}
\usepackage{stfloats}
\usepackage{url}
\usepackage{verbatim}
\usepackage{graphicx}
\usepackage{bbding}
\usepackage[dvipsnames]{xcolor}

\newlength \mywidth
\newlength \myw
\newlength \mywidthsupplora
\begin{document}

\title{NSYNC: Negative Synthetic Image Generation for Contrastive Training to Improve Stylized Text-To-Image Translation}

\author{Serkan~Ozturk, Samet~Hicsonmez\textsuperscript{\textdagger}, Pinar~Duygulu
\thanks{S. Ozturk, and P. Duygulu are at the Dept. of Computer Engineering, Hacettepe University, Ankara, Turkey. S. Hicsonmez is at the University of Luxembourg, Luxembourg. 
E-mail: {serkan.ozturk@hacettepe.edu.tr, samet.hicsonmez@uni.lu, pinar@cs.hacettepe.edu.tr}  \newline
\textsuperscript{\textdagger} Corresponding author.
}}

\markboth{}%
{NSYNC}


\maketitle

\begin{abstract}

Current text conditioned image generation methods output realistic looking images, but they fail to capture specific styles. Simply finetuning them on the target style datasets still struggles to grasp the style features. In this work, we present a novel contrastive learning framework to improve the stylization capability of large text-to-image diffusion models. Motivated by the astonishing advance in image generation models that makes synthetic data an intrinsic part of model training in various computer vision tasks, we exploit synthetic image generation in our approach. 
Usually, the generated synthetic data is dependent on the task, and most of the time it is used to enlarge the available real training dataset. With NSYNC, alternatively, we focus on generating negative synthetic sets to be used in a novel contrastive training scheme along with real positive images. In our proposed training setup, we forward negative data along with positive data and obtain negative and positive gradients, respectively. We then refine the positive gradient by subtracting its projection onto the negative gradient to get the orthogonal component, based on which the parameters are updated. This orthogonal component eliminates the trivial attributes that are present in both positive and negative data and directs the model towards capturing a more unique style.
Experiments on various styles of painters and illustrators show that our approach improves the performance over the baseline methods both quantitatively and qualitatively. 
Our code is available at \url{https://github.com/giddyyupp/NSYNC}.
\end{abstract}


\section{Introduction}
\label{sec:intro}

In recent years, we have witnessed significant progress in text-to-image translation, where the goal is to generate an image based on the content and the style elements described in the input text~\cite{reed}. 

Text to stylized image generation is an inherently difficult task where the model needs to generate both the content image and inject style components at the same time, especially compared to image-to-image translation based stylization where the input content image is already given.

Although first attempts~\cite{reed,stackgan,stackgan2,stylegant} were based on Generative Adversarial Networks, recently diffusion-based models~\cite{ddpm, ldm, diff_paperr} become the prominent approach for various image generation tasks including text-to-image generation. In these models, during training, latent vectors of input image and text are extracted from image and text encoders respectively, and then noise-added image latents are denoised using a text-conditioned denoising network.

Fine-grained image stylization is a specialized form of image stylization that focuses on transferring subtle, high-resolution stylistic attributes, such as brushstrokes and local texture variations, from a reference style image or text description to a target image. Unlike conventional image stylization, which typically applies global transformations involving overall color shifts, lighting changes, or broad texture patterns, fine-grained stylization preserves semantic structure while embedding localized artistic nuances that reflect the precise stylistic identity of the reference.

\begin{figure}[t]
\captionsetup[subfigure]{labelformat=empty}
\centering
\setlength\tabcolsep{1.5pt} 
\resizebox{1.0\columnwidth}{!}{
\begin{tabular}{@{}m{0.30\columnwidth}cccc:c}

{\textit{\textbf{~~~Input}}} & {\textit{\textbf{LDM~\cite{ldm}}}} & {\textit{\textbf{LoRA~\cite{lora}}}} & {\textit{\textbf{TI~\cite{ti_model}}}} & 
{\textit{\textbf{NSYNC}}}  & {\textit{\textbf{GT}}} \\
A serene pond with a white bridge, surrounded by autumn foliage and vibrant flowers. &
\includegraphics[width=0.16\textwidth,  ,valign=m, keepaspectratio,] {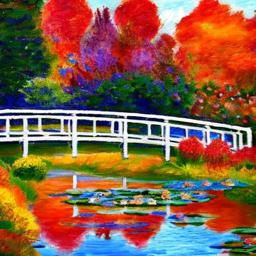} &
\includegraphics[width=0.16\textwidth,  ,valign=m, keepaspectratio,] {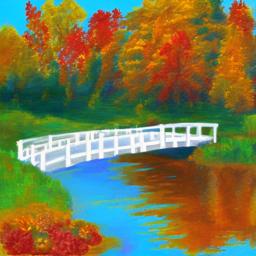} &
\includegraphics[width=0.16\textwidth,  ,valign=m, keepaspectratio,] {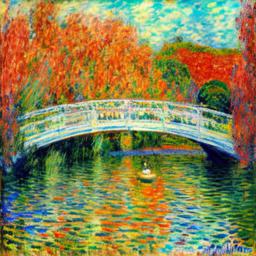} &
\includegraphics[width=0.16\textwidth,  ,valign=m, keepaspectratio,] {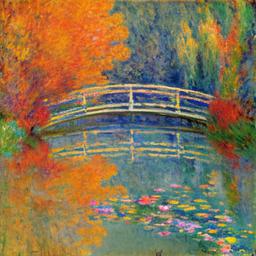} & 
\includegraphics[width=0.16\textwidth,  ,valign=m, keepaspectratio,] {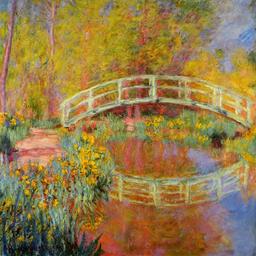} \\

The image features a vibrant night sky with swirling stars and a bright sun. A tree and a mountain range are visible below. &
\includegraphics[width=0.16\textwidth,  ,valign=m, keepaspectratio,] {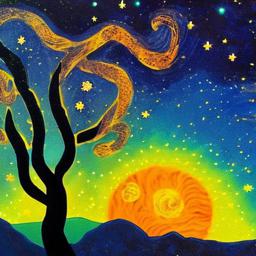} &
\includegraphics[width=0.16\textwidth,  ,valign=m, keepaspectratio,] {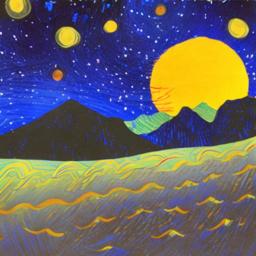}  & 
\includegraphics[width=0.16\textwidth,  ,valign=m, keepaspectratio,] {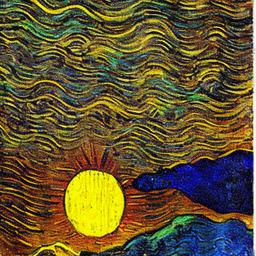} &
\includegraphics[width=0.16\textwidth,  ,valign=m, keepaspectratio,] {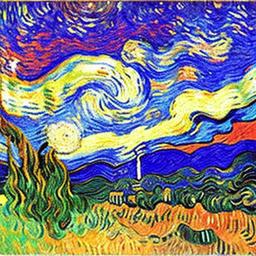} & 
\includegraphics[width=0.16\textwidth,  ,valign=m, keepaspectratio,] {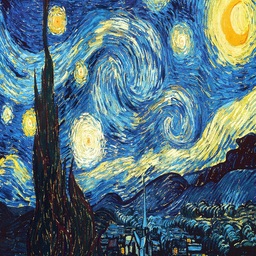}  \\

\end{tabular}}
\caption{Text-to-image generation results of an off-the-shelf Latent Diffusion Model (LDM)~\cite{ldm}, finetuned LoRA~\cite{lora} and Textual Inversion (TI) ~\cite{ti_model}, and our approach, NSYNC, on paintings of Monet at the top and Van Gogh at the bottom. It is clearly visible that LDM generates a generic image without the target style for the given input text. Although, TI~\cite{ti_model} improves over LDM, it still fails to capture style elements. On the other hand, our method NSYNC captures the target style and generates visually similar images to the real paintings. Note that for each method, we append the text \textit{in the style of $*$} to the input captions. Note that ground truth (GT) images denote the origin of the input text prompts. Zoom in for details.}
\label{fig:teaser}
\end{figure}

The models proposed for text-to-image translation achieve relatively low performance on \textit{fine-grained} stylized image generation tasks when used off-the-shelf (see Figure~\ref{fig:teaser}). Specifically, these models are successful at generating more general concepts such as ``impressionist paintings'' or ``image in the style of a children's illustration'', but fail at fine-grained styles like ``painting in the style of Monet'' or ``an illustration in Dr. Seuss style''.
This limitation likely occurs due to i) these large-scale models being trained on mostly natural images, 
and ii) the style being usually embedded in the text encoder at a broader level, i.e., \textit{Impressionist style} instead of \textit{the style of Monet}.

\renewcommand{\arraystretch}{3}

\renewcommand{\arraystretch}{1}

To finetune the baseline text-to-image diffusion models on specific sets of images or styles, two approaches have recently been proposed. LoRA~\cite{lora} focuses on updating the weights of the latent diffusion model by finding the low-rank adaptations instead of updating the entire network. Alternatively, Textual Inversion~\cite{ti_model} fixes the latent diffusion model, inserts a new token into the text encoder that describes the style or concept, and optimizes the embeddings of this new token, which eventually holds the target style. Even though these approaches improve the stylization performance, they fail to capture the nuances in the target style since they cannot sufficiently differentiate data belonging to the negative category that is visually similar to positives.

In this work, we propose a contrastive learning framework for fine-grained stylized text-to-image generation task.
Our goal is to enable the model to distinguish between trivial or generic style features and specific style features, thereby enhancing the fine-grained style learning process.

Utilization of contrastive loss~\cite{hadsell2006dimensionality} aims to maximize the intra-class similarity while maximizing the inter-class dissimilarity and has become an essential part of training self-supervised~\cite{contrastive_lr, cl_ref2}, semi-supervised~\cite{9732218} and fully-supervised~\cite{DBLP:conf/emnlp/ReimersG19,NEURIPS2020_d89a66c7} models for various tasks. 
On the other hand, it necessitates negative examples besides the positives.
Yet, gathering the required negative dataset through manual data collection and annotation is time consuming and cumbersome. Moreover, the collected negative set could introduce bias towards a specific concept or style.

Recently, synthetic images have become part of model training for various computer vision tasks, especially in image classification, object detection, and segmentation~\cite{he2023is, Sariyildiz_2023_CVPR, Tian_2024_CVPR, Singh_2024_CVPR, hammoud2024synthclip, karazija2024diffusion}. 
In addition to being used as supplementary data, synthetic datasets, even when used as the sole training data, allowed the models to achieve results on par with those trained exclusively on real datasets. Yet, synthetic data have been mostly used as a
positive set for model improvement.

In this work, we propose generating negative synthetic sets. For this purpose, we utilize current image generation models. Generating synthetic data is fast, and with the success of current models, the generated images hold the concepts presented in the input text very accurately. In addition, text-to-image models already know how to generate some generic features that are common across different concepts when learning a specific concept, such as style. Therefore, choosing negative data from synthetically generated images allows the model to focus on more distinctive traits.

In order to capture the unique styles, we further introduce a novel component in contrastive learning to eliminate trivial attributes that are shared both by positive and negative data.
We obtain positive and negative gradients by forwarding positive and negative data, respectively. Then, by subtracting the projection of the positive gradient onto the negative gradient from itself, we get the orthogonal gradient component and use this for parameter update.  

\begin{figure}
\centering
\includegraphics[width=1.0\columnwidth]{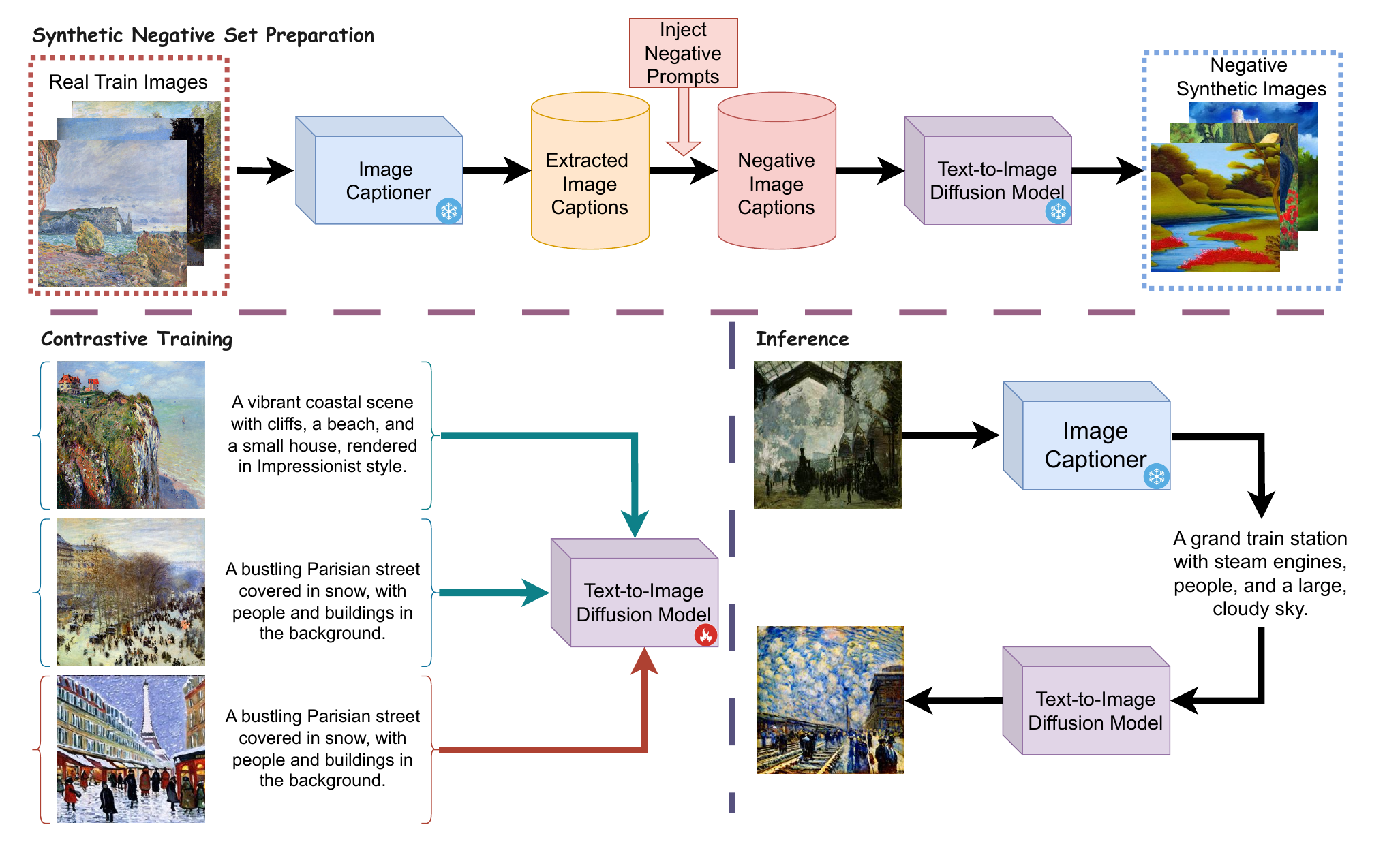}
\caption{NSYNC processing pipeline. We start with curating a negative dataset. Next, we finetune the baseline adapter model using our novel contrastive learning formulation. Finally, the inference stage is similar to baseline text-to-image diffusion model.}
\label{fig:overview}
\end{figure}

Overview of our framework is presented in Figure~\ref{fig:overview}. First, we generate synthetic images using an off-the-shelf latent diffusion model by feeding captions which capture the generic style, i.e. \textit{paintings}, \textit{impressionist painting} or \textit{illustrations}. Then, we finetune the baseline text-to-image generation model using positive and negative images with our novel contrastive learning approach to capture specific and fine-grained styles such as Monet and Van Gogh. Although any text-to-image generation method can be used as the finetune model, in this study, we utilize Textual Inversion~\cite{ti_model}. 

Let's consider a case where the goal is to create paintings similar to Monet. We first create captions for various natural images. Then using these text descriptions, we generate images with the style of not-Monet by adding a negative prompt to the text descriptions. In the final stage, we finetune the baseline text-to-image generation model by using Monet paintings as the positive set and not-Monet synthetic images as the negative set and employ a contrastive training approach.

Our contributions can be summarized as follows:
\begin{itemize}
    \item We attack the problem of \textit{fine-grained} stylized text-to-image generation task with a contrastive learning framework. 
    \item We propose to generate negative synthetic sets to improve the performance of the models.
    \item We propose a novel idea to use the gradient of the negative data to refine the gradient of the positive data, using orthogonal component.
    \item We show the effectiveness of our approach on capturing styles of various painters and illustrators.    
 \end{itemize}

\section{Related Work}
\label{sec:related}

\paragraph{Text-to-Image Generation}
In the short history of text-to-image generation literature, various approaches have been proposed.  
The seminal work of Reed et al.~\cite{reed} formulated the task as conditional image generation ~\cite{gan, stylegan, pix2pix, cyclegan, dualgan, huang2018multimodal, hicsonmez2021adversarial}. Input text descriptions are embedded and appended to the input latent vector, which is fed to a generator network to generate images, and a discriminator network is trained to distinguish generated images from real ones. 
Later, various GAN-based methods~\cite{stackgan, stackgan2} are built based upon this approach and achieved better generation results. 

Some other studies utilized the idea of auto-regressively generating each pixel of the image~\cite{dalle, Esser_2021_CVPR, yu2022vectorquantized}. Although good generation quality is achieved, outputs are limited to low resolution and huge computation power is demanded.

Recently, diffusion based models~\cite{diff_first, diff_paperr, ldm, saharia2022photorealistic} become dominant in the text-to-image generation task~\cite{dhariwal2021diffusion}. In Latent Diffusion Models, the input text is embedded in the denoising process using cross attention. 
The drawback of the diffusion based image generation models~\cite{diff_first, dhariwal2021diffusion, ddpm, ldm, diff_paperr} is that they usually require being trained on huge amounts of image data. Therefore, it is unlikely to achieve good performance by directly finetuning them on small datasets.

In order to solve this problem, 
data efficient adaptation or personalization to small datasets attracted attention~\cite{dreambooth, ti_model, lora, zhang2023adding, zhang2023sine, wang2024instantstyle, qi2024deadiff}. 
These adaptation models help to finetune a large baseline model using only a limited number of images, and achieve state-of-the-art performance on the target dataset. 
Among them, SINE~\cite{zhang2023sine} learns editing a single input image with text guidance. Here style is given using generic text prompts rather than learning a model using style datasets. On the other hand, InstantStyle~\cite{wang2024instantstyle} and DEADiff~\cite{qi2024deadiff} are driven by example style images. Content is generated based on given text prompts and stylization is done using the input style image. These approaches either rely on the stylization capabilities of the baseline diffusion models, or require feeding an example style image. However, our goal is to learn the specific style cues inherent in given art images, rather than using a limited number of images exhaustively. 

Textual Inversion~\cite{ti_model} aims to train a special token extracted from a small set of images that is common to all given images and uses this token to sample from the baseline stable diffusion model. LoRA~\cite{lora}, which was originally proposed to efficiently finetune Large Language Models, is adapted to image generation as well. LoRA adapts the baseline model to a set of images by learning low ranked matrices that approximate full model updates. Both approaches provide a suitable solution for our goal.
In this work, we adopt Textual Inversion~\cite{ti_model} as the primary fine-tuning approach, and further integrate our method with LoRA~\cite{lora} to demonstrate the generalization capability of NSYNC across different adaptation paradigms.

\paragraph{Contrastive Learning}
Contrastive loss~\cite{hadsell2006dimensionality} aims to reduce the distance between the representations of the data from the same class while increasing the distance to other classes. Contrastive loss has been effective in various computer vision tasks, especially for unsupervised or self-supervised representation learning \cite{contrastive_lr, cl_ref2, He_2020_CVPR, rep_learn}, and is also utilized in image-to-image translation and conditional image generation \cite{cut, wang2021instance, zhan2022marginal}. 

CUT~\cite{cut} is one of the first studies that used contrastive training for image-to-image translation. Patches are extracted from input, and images are translated by grouping patches from the same location as positive and others as negative. 
Wang et al.~\cite{wang2021instance} improved the negative sampling of CUT, aiming to generate hard negatives. Patch based approach is then extended to guided image generation~\cite{zhan2022marginal}.

Whether it is image or patch based, selection of positive and negative pairs has a direct effect on performance. Especially when there is no labeled negative set, this selection becomes an important issue. Currently, the most dominant approach is to use augmentations on the image level~\cite{contrastive_lr}, by using patches from the same image as positive and patches from other images as negative~\cite{dino}. In this work, we propose to use synthetically generated samples as the negative set.

\paragraph{Synthetic Data}
The success of image generation methods has attracted a lot of attention to use synthetic data generated by text-to-image models in training models for different vision tasks. Various works~\cite{he2023is, azizi2023synthetic, Sariyildiz_2023_CVPR, Tian_2024_CVPR, Singh_2024_CVPR, hammoud2024synthclip, Lin_2023_CVPR} showed the usability of synthetic data and compared its performance to using real data on different tasks and datasets. \cite{he2023is} generated synthetic data to finetune CLIP~\cite{clip} and evaluate its few shot capabilities on 17 image classification datasets.~\cite{hammoud2024synthclip} showed comparable performance by training CLIP from scratch using only synthetic data. \cite{Sariyildiz_2023_CVPR} focused on ImageNet~\cite{deng2009imagenet} and compared the performance of models trained on only synthetic data to current benchmarks. \cite{Lin_2023_CVPR} utilized text-to-image generated synthetic data for Few Shot Object Detection (FSOD) task and showed that additional synthetic data outperforms baseline models.

In all these works, synthetic data is used to either augment or replace the real dataset. In our work, we exclusively generate synthetic data that is complementary to our real dataset, and use it in a contrastive learning framework.

\section{Method}
\label{sec:method}

\begin{figure*}
\centering
\includegraphics[width=1.0\textwidth]{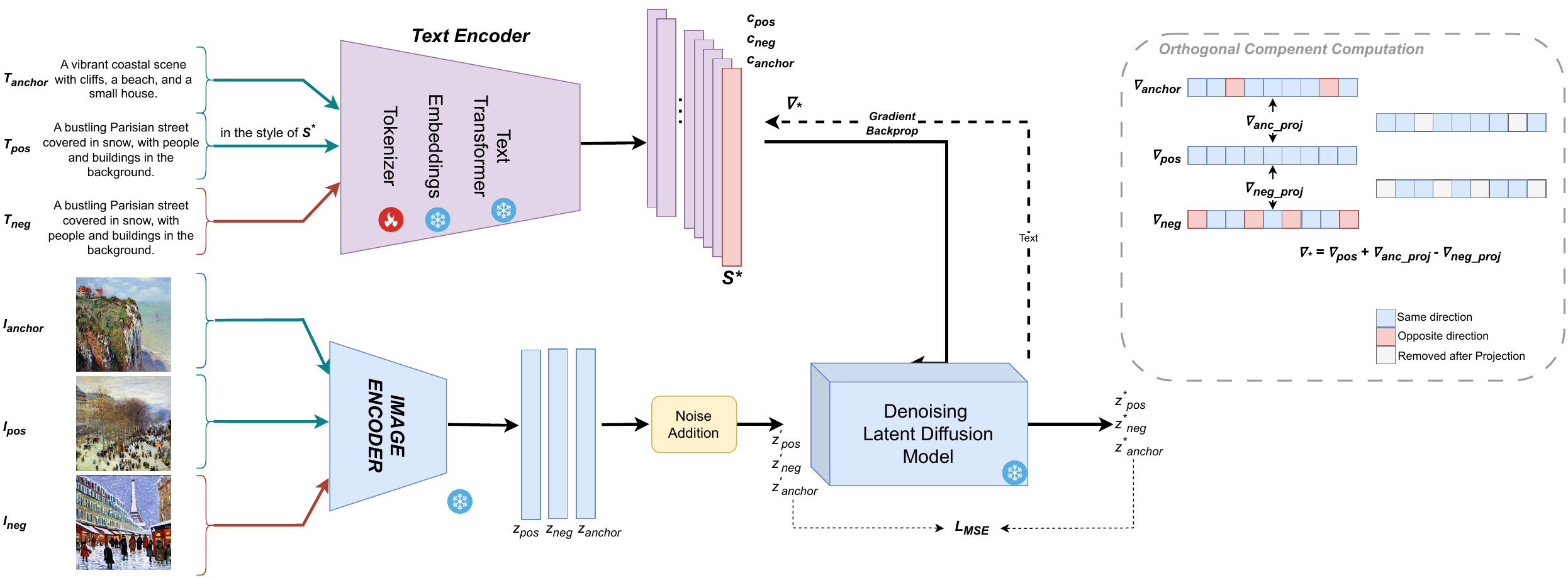}
\caption{Contrastive training framework of NSYNC. We only train the embedding of the special style token in the Text Encoder similar to Textual Inversion. However, instead of directly updating the weights ($v_{*}$) of the newly added token $S^*$ using only a single positive image, we calculate two projections on the gradients of positive (${\nabla}_{pos}$), negative (${\nabla}_{neg}$) and anchor (${\nabla}_{anchor}$) images. We use these projections to better find the gradient directions, and update the weight $v_{*}$ with the refined gradient ${\nabla}_{*}$
} 
\label{fig:model}
\vspace{-3mm}
\end{figure*}

We present the overview of our framework in Figure~\ref{fig:overview} and the training process in Figure~\ref{fig:model}. Our goal is to train a text-to-image generation model using a small amount of data consisting of only images in the target style, and the text descriptions that are generated using an off-the-shelf model.
Our method could be separated into two sub-components: negative set curation and contrastive training framework, which utilizes Textual Inversion as the adaptation model for finetuning. 
The details will be provided in the following.

\paragraph{Negative Set Creation} 
We start by curating a negative/generic image dataset by making use of a state-of-the-art latent diffusion model (LDM). 
First, we extract detailed captions for the training images using InternVL~\cite{captioner} model. Then, we add a negative prompt depending on the style category (e.g. painting, illustration) to all these captions, and forward these text descriptions to a frozen text-to-image generation model~\cite{ldm}. The SD model is capable of generating artistic images with generic concepts. Here, our goal is to utilize the \textit{negative} images that hold only generic styles, to refine the training of a specific style.

\paragraph{Textual Inversion~\cite{ti_model}} Textual Inversion (TI) aims to learn a new token embedding for the target style or concept in general. They use a pre-trained text-to-image generation latent diffusion model where they froze the weights of the variational auto-encoder (VAE) and diffusion UNet models. They append a new token ($S^*$) to the text encoder and update only the embedding weights of this new token. They append this new token to the captions of each image and forward the image and text from the image and text encoders, respectively. Then the encoded image is contaminated with noise and fed to the denoising step. During denoising, noised image latent is conditioned with text embeddings using cross attention. Finally, the mean squared error loss is calculated between the added and predicted noises. Here, since the VAE and UNet models are fixed, only the embedding of the new token is updated by zeroing the gradients of the present tokens. The loss function used in Textual Inversion is defined as:

\begin{equation}
\label{ti_loss}
\resizebox{0.9\columnwidth}{!}{%
$v_* = \arg\min_v \, \mathbb{E}_{z \sim \mathcal{E}(x), \, y, \, \epsilon \sim \mathcal{N}(0, 1), \, t} \left[ \left\| \epsilon - \epsilon_\theta \left( z_t, t, c_\theta(y) \right) \right\|_2^2 \right] $%
}
\end{equation}

\noindent where $\mathcal{E}$ is the VAE that maps the input image $x$ to latent space, $\epsilon_\theta$ is the denoising network, $c_\theta$ is the text encoder with the additional token, $y$ is the text condition, and $z_t$ is the noised latent over $t$ time steps.

\paragraph{Contrastive Training} At this stage, we have a positive dataset composed of our target style images and their text descriptions, and a negative dataset generated in the previous step by using the same text descriptions with only negative/generic style prompts. 
We finetune the baseline TI~\cite{ti_model} model by employing our proposed novel contrastive training framework.

During training, we sample one negative (neg) image and one positive (pos) image and use the same text description for both of them, i.e. no additional negative prompt injection. We also sample another positive image named as \textit{anchor}. We follow the same forward step as in TI~\cite{ti_model}, but instead of directly updating the embedding of the new token, we first accumulate the gradients coming from pos, anchor, and neg images. Then, we calculate the projection of the gradient coming from the pos image onto the neg one, and subtract this projection from the gradient of the pos image. Similarly, we repeat the same process using pos and anchor images, and this time we add the resulting projection to the gradient of the pos image. Our intuition is that, if we remove the components that exist in both pos and neg from the pos gradient, then we could direct the pos gradient in a direction that diverges from the neg gradient the most, thus leading us to more unique features in the target style. This process is formulated as:

\begin{equation}
\label{our_loss}
\begin{aligned}
  {\nabla}_{neg-{proj}}  &= \frac{{\nabla}_{pos} \cdot {\nabla}_{neg}}{{{{\nabla}_{neg}}}^2} {\nabla}_{neg} \\
    {\nabla}_{anc-{proj}}  &= \frac{{\nabla}_{pos} \cdot {\nabla}_{anc}}{{{{\nabla}_{anc}}}^2} {\nabla}_{anc} \\
  {\nabla}_{*} &= {\nabla}_{pos} - {\nabla}_{{neg-}proj} + {\nabla}_{anc-{proj}}
\end{aligned}
\end{equation}

\noindent where ${\nabla}_{pos}$, ${\nabla}_{anc}$ and ${\nabla}_{neg}$ are the gradients of the textual inversion loss coming from positive, anchor, and negative images, respectively. 
This process is presented in Figure~\ref{fig:model}.

\paragraph{Inference} During inference, for a given image, we first extract the caption and append the special token. Then, encode the text using the updated text encoder and forward it to the diffusion model to generate the image.  

\begin{table}[th]
\centering
\resizebox{1.0\columnwidth}{!}{
\begin{tabular}{l|cc|c|cc|c}
\toprule
 & M & VG &  SG & PP & MB & PM    \\
 \midrule
 \# Positives & 1072 & 400  & 500 & 447  & 272 & 137  \\
 \# Test &  121 & 400 & 311 & 309 & 189 & 39   \\
\bottomrule
\end{tabular}}
\caption{Statistics of the datasets used. Paintings: M-Monet, VG-Van Gogh; Animation: SG-Studio Ghibli; Illustrations: PP-Patricia Polacco, MB-Marc Brown; Abstract Art: PM-Piet Mondrian.}
\label{tab:dataset}
\vspace{-3mm}
\end{table}
\section{Experiments}
\label{sec:exp}

\begin{table*}
\begin{center}
\resizebox{1.0\textwidth}{!}{
\begin{tabular}{lccccc:cccc:cccc}
 \toprule 
 \multirow{3}{*}{Method}  & \multirow{3}{*}{SD} & \multicolumn{4}{c}{Monet} & \multicolumn{4}{c}{Van~Gogh} & \multicolumn{4}{c}{Studio Ghibli}  \\
  \cmidrule(lr){3-6}  \cmidrule(lr){7-10}  \cmidrule(lr){11-14} 
& & CSD$\uparrow$ & CMMD$\downarrow$  & KID$\downarrow$ & FID$\downarrow$ & CSD$\uparrow$ & CMMD$\downarrow$  & KID$\downarrow$ & FID$\downarrow$ & CSD$\uparrow$ & CMMD$\downarrow$  & KID$\downarrow$ & FID$\downarrow$   \\
\midrule 
InST~\cite{wang2024instantstyle} & 1.4 & 0.6347 & 1.382 & 0.0748 & 198.8 & 0.7689 & 0.778 & 0.0532 & 134.4& 0.6364 & 1.122 & 0.0148 & 121.8\\
StyleShot~\cite{gao2024styleshot} & 1.5 & 0.3550 & 3.728 & 0.0515 & 175.6 & 0.4363 & 2.038& 0.0542 & 144.4& 0.4686 & 1.688 & 0.0356 & 139.1 \\
StyleShot$\dagger$~\cite{gao2024styleshot} & 1.5 & 0.3715 & 2.576 & 0.0334 & 167.4& 0.4567 & 2.088& \textbf{0.0233} & \textbf{117.2} & 0.3827 & 2.207& 0.0126 & 125.9\\
DEADiff$\dagger$~\cite{qi2024deadiff} & 1.5 & 0.1199 & 3.402 & 0.0642 & 198.3& 0.3275 & 2.567& 0.0674 & 169.5&  0.5434 & 1.478& 0.0282 & 128.7\\
SD$\dagger$~\cite{ldm} & 1.5 & 0.6089 & 1.016 & \underline{0.0257} & 156.5 & 0.7172 & 1.299& 0.0462 & 136.9 & 0.4456 & 1.149 & 0.0210 & 131.7 \\
TI~\cite{ti_model} & 1.5 & 0.6182 & 1.034 & 0.0396 & 171.4 & \underline{0.8026} & 0.800& 0.0586 & 143.6 & 0.6594 & \textbf{0.749} & 0.0146 & 121.2 \\
SD$\dagger$~\cite{ldm} & 2.0 & 0.3022 & 1.996 & 0.0363 & 165.8& 0.5182 & 1.190 & 0.0410  & 139.6 & 0.3834 & 1.559& 0.0257 & 128.7\\
TI~\cite{ti_model} & 2.0 & 0.6374 & 0.964 & 0.0443 & 163.9 & 0.7604 & \underline{0.583} & 0.0613 & 136.1 & 0.7188 & 0.805& \textbf{0.0091} & \textbf{113.4}\\
\midrule
NYSNC & 1.5 & \textbf{0.7232} & \underline{0.957} & 0.0266 &  \underline{146.5} & \textbf{0.8098} & \underline{0.746} & \underline{0.0388} & \underline{125.6} & \underline{0.7420} & \underline{0.764} & \underline{0.0125} & 121.1 \\
NYSNC & 2.0 & \underline{0.6484} & \textbf{0.721} & \textbf{0.0232}  & \textbf{139.5} & 0.7905 & \textbf{0.518} & 0.0600 & 135.1 & \textbf{0.7497} & 0.853& 0.0241 & \underline{118.1} \\
 \bottomrule 
\end{tabular}}
\end{center}
\caption{Results on the painters and animation datasets. Best results are boldfaced, 2nd best are underlined. $\dagger$ results with pretrained model. Our method achieves the top-2 scores on all metrics.}
\label{tab:painter_res}
\vspace{-3mm}
\end{table*}

\subsection{Datasets}
In our experiments, we focused on the art domain and selected four different categories; paintings, animation movies, children's book illustrations, and abstract art. For the painters (Monet and Van Gogh), we used the same train and test sets proposed in~\cite{cyclegan}; for the animation style (Studio Ghibli), we used a publicly available dataset in HuggingFace (Nechintosh/ghibli); for the children's illustrations, we chose two illustrators (Patricia Polacco and Marc Brown) from the illustration dataset presented in~\cite{ganilla}; and finally, for the abstract art, we used images of a famous artist (Piet Mondrian) from a publicly available dataset~\cite{gontier2023delaunay}. 

In order to create the negative datasets, we first extract the captions from the training images and then forward them to a pretrained Latent Diffusion Model~\cite{ldm} model to generate the images. For all datasets, we set the number of negative images to be the same as the number of positive images. The only difference is the addition of a negative prompt, which is adapted based on the target style. 
Table~\ref{tab:dataset} shows the number of positive images (respectively, negative images) in the train set and the number of images in the test set for all datasets.

\subsection{Implementation Details} We used PyTorch~\cite{pytorch} to implement our models. All training images are resized to $512\times512$ pixels. Textual Inversion~\cite{ti_model} is used as the adaptation method. The model is trained for $8000$ iterations on all datasets, using the Adam~\cite{adam} solver with a learning rate of 0.0008 and a batch size of 8. We conducted all our experiments on Nvidia Tesla A100 GPUs. For the baseline methods, we used publicly available official repositories. During inference, we use the same number of DDIM sampling steps ($50$) for all methods.

\subsection{Metrics}

For image generation, Fréchet Inception Distance (FID)~\cite{FID} and Kernel Inception Distance (KID)~\cite{KID} are the two most commonly used metrics to evaluate visual similarity between real and generated images. 
Both metrics rely on an Inception~\cite{szegedy2016rethinking} model trained on Imagenet~\cite{deng2009imagenet} to extract image features from both real and generated images to calculate the distance between them.
The distinction lies in the calculation of the difference. FID uses the Fréchet distance, whereas KID uses Maximum Mean Discrepancy (MMD). One advantage of KID is that it provides reliable results with a small number of images compared to FID.

However, the current text-to-image generation models are capable of generating complex scenes, and the relatively weak image features coming from an Inception model struggle to capture the details in these images. Recently, CMMD~\cite{CMMD} has been proposed as an alternative metric. It uses rich CLIP~\cite{clip} image embeddings extracted from both real and generated images and calculates the Maximum Mean Discrepancy (MMD) of the two sets to measure the distance. It could be considered as KID with CLIP image embeddings. When the scene contains complex elements, such as paintings, this metric is more reliable. 

Contrastive Style Descriptors (CSD)~\cite{csd} is presented to measure style similarity in diffusion models. Following the procedure in~\cite{csd}, we compute the CSD score by: (1) calculating a mean style feature vector from all test set images and (2) measuring the cosine similarity between each generated image and this mean vector. We report the average similarity as the final score.

We opted not to conduct a user study, primarily due to the difficulty in recruiting a sufficient number of qualified evaluators with domain expertise in visual aesthetics. Moreover, such studies tend to introduce subjective biases, which we aimed to avoid in favour of more reproducible and objective evaluations.

\begin{table*}
\begin{center}
\resizebox{1.0\textwidth}{!}{
\begin{tabular}{lccccc:cccc:cccc}
 \toprule 
 \multirow{3}{*}{Method}  & \multirow{3}{*}{SD} & \multicolumn{4}{c}{PP} & \multicolumn{4}{c}{MB} & \multicolumn{4}{c}{PM}  \\
  \cmidrule(lr){3-6}  \cmidrule(lr){7-10}  \cmidrule(lr){11-14} 
& & CSD$\uparrow$ & CMMD$\downarrow$  & KID$\downarrow$ & FID$\downarrow$ & CSD$\uparrow$ & CMMD$\downarrow$  & KID$\downarrow$ & FID$\downarrow$ & CSD$\uparrow$ & CMMD$\downarrow$  & KID$\downarrow$ & FID$\downarrow$   \\
\midrule 
InST~\cite{wang2024instantstyle} & 1.4              & 0.5394 & 0.746 & 0.0442 & 152.6            & 0.4977 & 2.947 & 0.0618 & 138.1                          & 0.5035 & 1.730 & 0.0733 & 245.9 \\
StyleShot~\cite{gao2024styleshot} & 1.5             & 0.4829 & 1.052 & 0.0408 & 148.9            & 0.5327  & 1.892 & 0.0387 & 104.7                          & 0.5673 & 1.664 & 0.0816 & 234.1 \\
StyleShot$\dagger$~\cite{gao2024styleshot} & 1.5    & 0.3823 & 0.827 & 0.0148 & \textbf{138.5}   & 0.5147 & 1.767 & \textbf{0.0167} & 101.9                      & 0.6073 & 1.465 & 0.0587 & 205.7 \\
DEADiff$\dagger$~\cite{qi2024deadiff} & 1.5         & 0.3404 & 1.077 & 0.0386 & 158.5            & 0.4989 & 2.077 & 0.0533 & 119.7                          & 0.5989 & 1.951 & 0.1755 & 275.5 \\
SD$\dagger$~\cite{ldm} & 1.5                        & 0.3199 & 0.803 & 0.0384 & 161.6            & 0.4511 & 1.554 & 0.0363 & 136.2                         & 0.6773 & 1.137 & 0.0957 & 226.2 \\
TI~\cite{ti_model} & 1.5             & \underline{0.5562} & \textbf{0.669} & 0.0562 & 154.8      & 0.6114 & \underline{1.236} & 0.0429 & 120.5                  & 0.6189 & 0.979 & 0.0602 & 198.9 \\
SD$\dagger$~\cite{ldm} & 2.0                        & 0.3340 & 1.759 & 0.0659 & 168.4            & 0.4021 & 2.663 & 0.0428 & 116.5                  & \underline{0.7019} & \underline{0.704} & 0.0582 & 179.4 \\
TI~\cite{ti_model} & 2.0                            & 0.4929 & 1.557 & 0.0437 & 146.4            & \underline{0.6183} & 1.397 & 0.0332 & \underline{94.3}                  & 0.6357 & 0.925 & \underline{0.0489} & 183.1 \\
\midrule
NYSNC & 1.5     & \textbf{0.5840} & \underline{0.790} & \textbf{0.0344} & \underline{143.0}      & \textbf{0.6759} & \textbf{1.174} & \underline{0.0302} & \textbf{94.2}        & 0.6833 & 0.912 & \textbf{0.0421} & \textbf{169.1} \\
NYSNC & 2.0     & 0.4953 & 1.588 & 0.0606 & 151.6                   & 0.6162 & 1.532 & 0.0343 & 97.8                        & \textbf{0.7235} & \textbf{0.491} & 0.0744 & \underline{170.2} \\
 \bottomrule 
\end{tabular}}
\end{center}
\caption{Results on the illustration and abstract art datasets. Best results are boldfaced, 2nd best are underlined. $\dagger$ results with pretrained model. Our method achieves the best scores on CSD metric on all datasets.}
\label{tab:ill_abs_res}
\vspace{-4mm}
\end{table*}

\subsection{Baselines}
\label{sec:baselines}

There are a few works that are generating fine-grained style images directly from text inputs. Stable Diffusion (SD)~\cite{ldm} and Textual Inversion (TI)~\cite{ti_model} are the natural baselines with the same text-to-image stylization pipeline as ours. To enable a broader comparison, we also include recent diffusion-based image-guided stylization methods: InST~\cite{wang2024instantstyle}, StyleShot~\cite{gao2024styleshot}, and DEADiff~\cite{qi2024deadiff}, which achieve state-of-the-art results for stylization using image modality. While these methods differ fundamentally in their design and input requirements, typically relying on either a style image and a target prompt (DEADiff~\cite{qi2024deadiff}) or both style and content images (InST~\cite{wang2024instantstyle}, StyleShot~\cite{gao2024styleshot}), our approach performs stylization purely from a text prompt, without requiring any guiding image. Thus, a direct comparison is inherently asymmetric. Nevertheless, we designed a controlled experimental setting to enable a meaningful comparison.

For InST and StyleShot, which require both style and content images, we generate content images, similar to the negative set creation, using the same prompts as in our method (with the “in the style of *” clause removed). Style images are randomly sampled from the test set. For DEADiff, which requires a style image and a prompt, we use the same randomly selected style image and the same prompt used to generate the content image above (without modification). This ensures that all three methods are evaluated using consistent content and style inputs derived from the same source prompts and images.

Although DEADiff’s training code is unavailable, their released pretrained model includes Monet, Van Gogh, and Studio Ghibli styles, which we use directly for inference. StyleShot provides both training code and pretrained models that include these styles; we report results using both the released model and a version fine-tuned on our datasets. InST, which does not offer pretrained models, is trained from scratch using the official repository and our datasets.

\subsection{Quantitative Results}

Results on paintings and animation datasets are presented in Table~\ref{tab:painter_res}. 
Our method consistently achieves the best or second-best performance compared to all baseline approaches across all metrics. This shows that 
the proposed model better captures the fine-grained style of the target dataset. Especially on the Monet dataset, where we have more training images, NSYNC improves the baselines by a large margin across all metrics. The Van Gogh dataset contains many portraits and still-life paintings of fruits and flowers, where the scenes are relatively simple. That could be one of the reasons why the SD baseline achieves such good performance, especially on Inception based metrics. Except for InST~\cite{wang2024instantstyle}, the strong baselines StyleShot~\cite{gao2024styleshot} and DEADiff~\cite{qi2024deadiff} achieve sub-par performance, which is even lower than that of the baseline SD model. Finetuning the StyleShot method on the target datasets failed to bring any improvements, except for Studio Ghibli, where finetuning significantly increased the CSD and CMMD metrics.

On the illustrations and abstract art datasets, although the scenes are relatively simple compared to paintings and animations, i.e., composed of a simple and limited number of foreground and background objects, the style variations are more severe, and the specific styles do not exist in the baseline text-to-image generation model. On these datasets, see Table~\ref{tab:ill_abs_res}; our method again achieves the top-2 performances on all metrics. CMMD scores are similar to or better than the results of TI~\cite{ti_model}. On such simple images with few objects, the KID and FID metrics could better represent the visual similarity between generated and real images.

We also conduct an experiment in which we train two styles simultaneously in a single model by adding two separate tokens for each style. We use the Monet and Van Gogh datasets for this experiment, and the quantitative metrics are similar to those in the separate model trainings; see Table~\ref{tab:painter_res}. On Van Gogh, FID and KID metrics are improved to $129.8$ and $0.049$, respectively, while on Monet, FID increases by 5 points and KID remains the same.

In order to demonstrate the generalisation capability of our method, we extend our implementation to LoRA~\cite{lora} based finetuning and train models on all the datasets. We use Stable Diffusion version 2.0 for LoRA experiments. In Table~\ref{tab:all_lora_res}, we present a comparison between plain LoRA training using the dataset and our method, NSYNC. 
Overall, LoRA fails to capture the fine-grained style details, which is reflected in the low CSD and high CMMD scores compared to the TI based results. Nonetheless, in most cases, our method improves the CSD and CMMD metrics, which shows that NSYNC is not bound to a specific training scheme and could be integrated into different approaches.

\begin{table*}
\begin{center}
\begin{tabular}{l cc:cc:cc:cc:cc:cc}
 \toprule 
 \multirow{3}{*}{Metric}  & \multicolumn{2}{c}{Monet} & \multicolumn{2}{c}{Van Gogh}  & \multicolumn{2}{c}{Studio Ghibli} & \multicolumn{2}{c}{PP} & \multicolumn{2}{c}{MB} & \multicolumn{2}{c}{PM}      \\
  \cmidrule(lr){2-3}  \cmidrule(lr){4-5}  \cmidrule(lr){6-7}  \cmidrule(lr){8-9} \cmidrule(lr){10-11} \cmidrule(lr){12-13}
 & LoRA & NSYNC   & LoRA & NSYNC & LoRA & NSYNC &  LoRA & NSYNC &  LoRA & NSYNC &  LoRA & NSYNC\\
 
\midrule 
CSD$\uparrow$  &   0.1478   &  \textbf{0.1977} &   0.3820  & \textbf{0.4021}  & 0.2129 & \textbf{0.2152}   &  0.3573 & \textbf{0.3728} & 0.4490 & \textbf{0.4940} & \textbf{0.5281} & 0.5083    \\

CMMD$\downarrow$  &  \textbf{2.819}   &   2.870   &  1.974  & \textbf{1.800}   & 3.348 & \textbf{3.289}  &   1.494 & \textbf{1.394}  & 2.430  & \textbf{2.355}  & 1.601 &  \textbf{1.433}  \\

KID$\downarrow$ & 0.045  & \textbf{0.038}  &  \textbf{0.019} &  0.021 & \textbf{0.033} & 0.042 &   0.041 & \textbf{0.027}  & 0.033  & \textbf{0.031} & \textbf{0.065} &  0.077 \\

FID$\downarrow$ & 176.6 & \textbf{161.2}  & \textbf{116.5} & 120.2  & \textbf{146.6} & 158.6 &    151.2 & \textbf{149.4}  &  113.1 & \textbf{106.9}& 236.8 & \textbf{214.0}  \\ 
 \bottomrule 
\end{tabular}
\end{center}
\caption{Results on the illustration dataset using LoRA~\cite{lora} as the finetuning method. Best results are boldfaced. Our method improves the baseline on CSD and CMMD metrics in most occasions.}
\label{tab:all_lora_res}
\end{table*}

\renewcommand{\arraystretch}{1}
\begin{figure*}
\captionsetup[subfigure]{labelformat=empty}
\centering
\setlength\tabcolsep{1.5pt} 
\resizebox{1.0\textwidth}{!}{
\begin{tabular}{ccccccc}

{\textit{\textbf{\tiny{}}}} & {\textit{\textbf{\tiny{M}}}} & {\textit{\textbf{\tiny{VG}}}} & {\textit{\textbf{\tiny{SG}}}} &  {\textit{\textbf{\tiny{PP}}}} & {\textit{\textbf{\tiny{MB}}}} &  {\textit{\textbf{\tiny{PM}}}} \\

{\rotatebox[origin=b]{90}{\textit{\textbf{\tiny Desc.}}}} 
& \multicolumn{1}{p{1.5cm}}{\tiny{A serene pond with blooming lilies and vibrant flowers, surrounded by lush greenery.}} 
& \multicolumn{1}{p{1.5cm}}{\tiny{A cobblestone street at night, lit by warm, golden lights from a café, with a starry sky overhead.}}
& \multicolumn{1}{p{1.5cm}}{\tiny{A serene painting scene with a easel, umbrella, and lush greenery, under a clear blue sky.}}
& \multicolumn{1}{p{1.5cm}}{\tiny{A woman in a patterned headscarf and apron is washing clothes in a rustic kitchen.}}
& \multicolumn{1}{p{1.5cm}}{\tiny{Three mice in a cozy room, one wearing glasses and a green sweater, one a pink dress, and a blue and white striped shirt.}}
& \multicolumn{1}{p{1.5cm}}{\tiny{A grid of squares in various colors, including red, yellow, blue, and black.}} \\

{\rotatebox[origin=t]{90}{\textit{\textbf{\tiny SD}}}} & 
\includegraphics[width=\mywidth,  ,valign=m, keepaspectratio,] {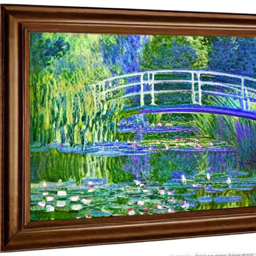} &
\includegraphics[width=\mywidth,  ,valign=m, keepaspectratio,] {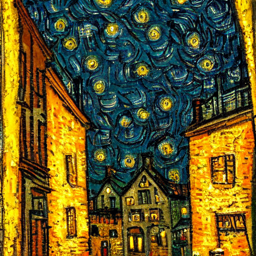} &
\includegraphics[width=\mywidth,  ,valign=m, keepaspectratio,] {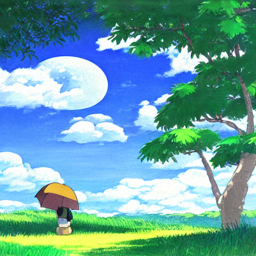} & 
\includegraphics[width=\mywidth,  ,valign=m, keepaspectratio,] {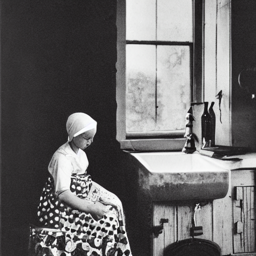} & 
\includegraphics[width=\mywidth,  ,valign=m, keepaspectratio,] {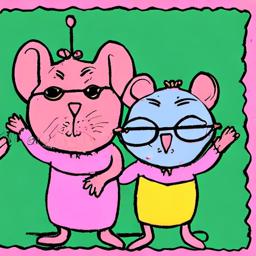} & 
\includegraphics[width=\mywidth,  ,valign=m, keepaspectratio,] {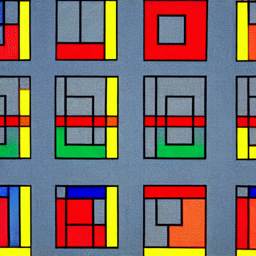} \\

{\rotatebox[origin=t]{90}{\textit{\textbf{\tiny InST}}}} & 
\includegraphics[width=\mywidth,  ,valign=m, keepaspectratio,] {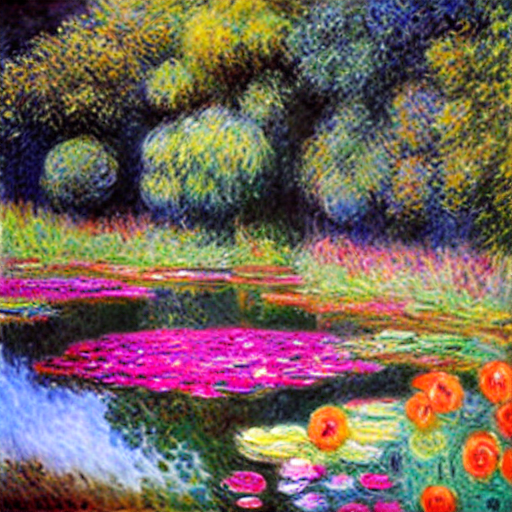} &
\includegraphics[width=\mywidth,  ,valign=m, keepaspectratio,] {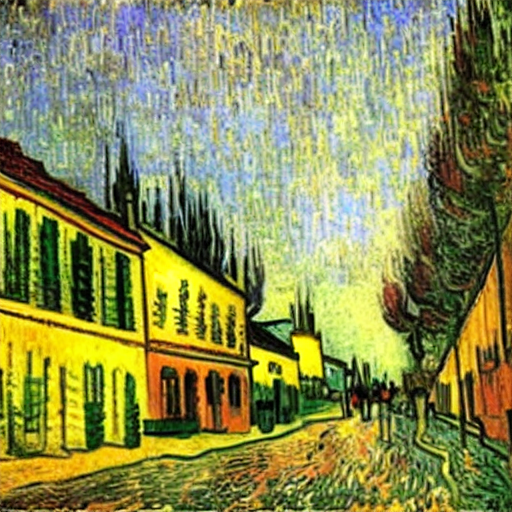} &
\includegraphics[width=\mywidth,  ,valign=m, keepaspectratio,] {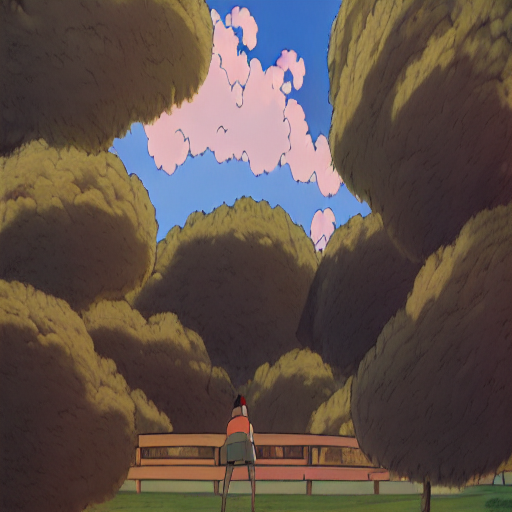} & 
\includegraphics[width=\mywidth,  ,valign=m, keepaspectratio,] {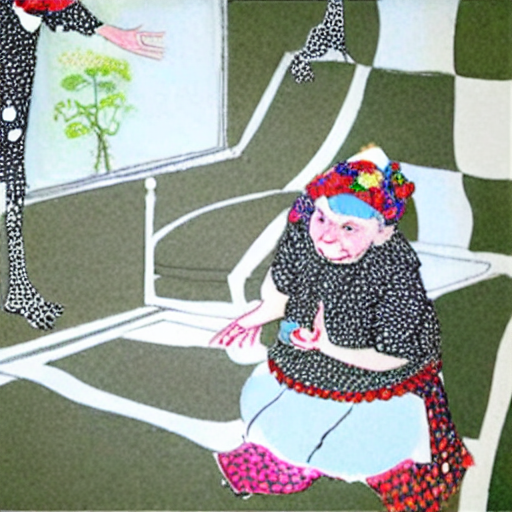} & 
\includegraphics[width=\mywidth,  ,valign=m, keepaspectratio,] {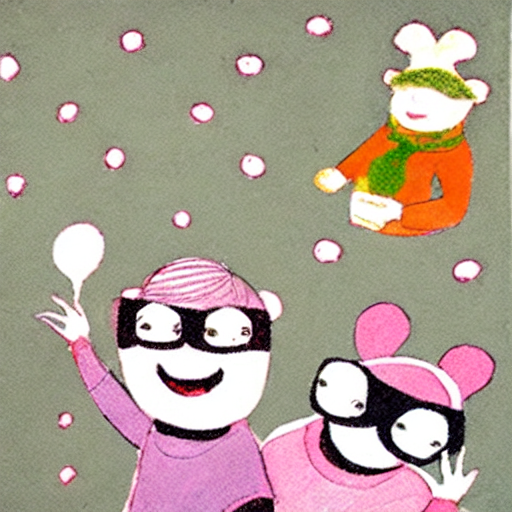} & 
\includegraphics[width=\mywidth,  ,valign=m, keepaspectratio,] {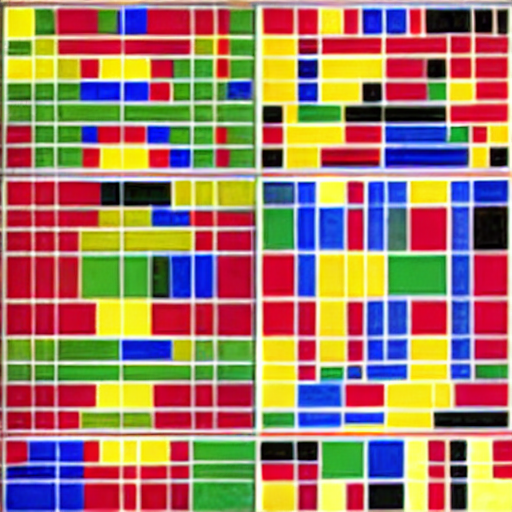} \\

{\rotatebox[origin=t]{90}{\textit{\textbf{\tiny StyleShot}}}} & 
\includegraphics[width=\mywidth,  ,valign=m, keepaspectratio,] {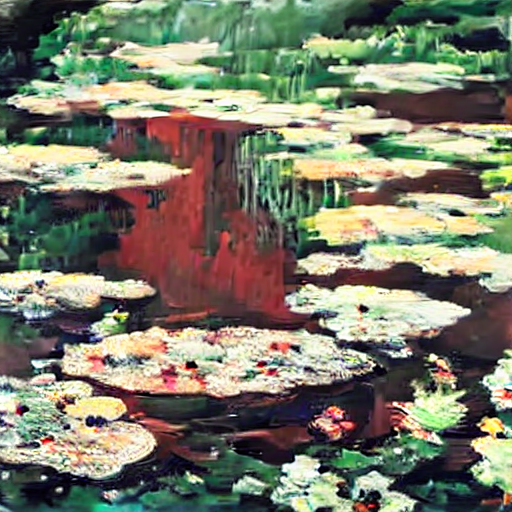} &
\includegraphics[width=\mywidth,  ,valign=m, keepaspectratio,] {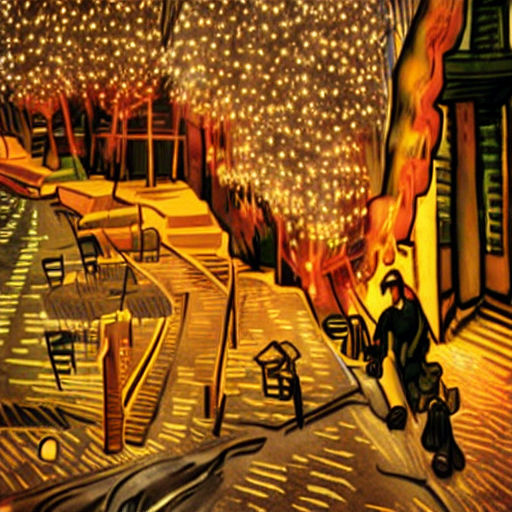} &
\includegraphics[width=\mywidth,  ,valign=m, keepaspectratio,] {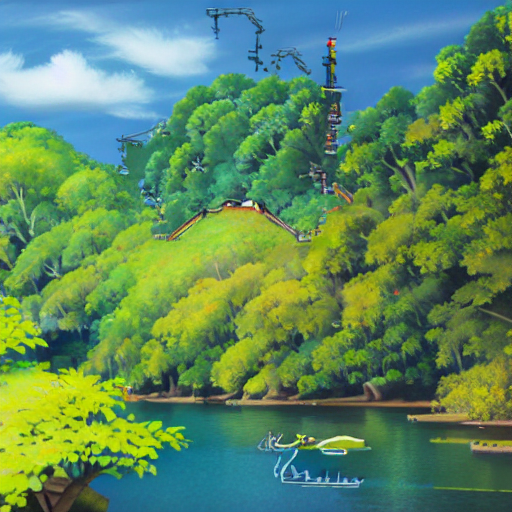} & 
\includegraphics[width=\mywidth,  ,valign=m, keepaspectratio,] {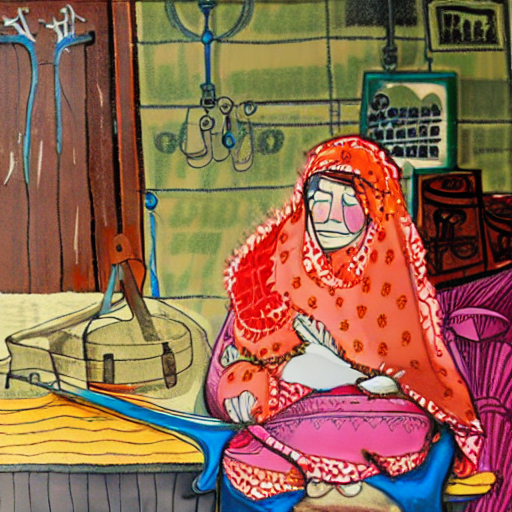} & 
\includegraphics[width=\mywidth,  ,valign=m, keepaspectratio,] {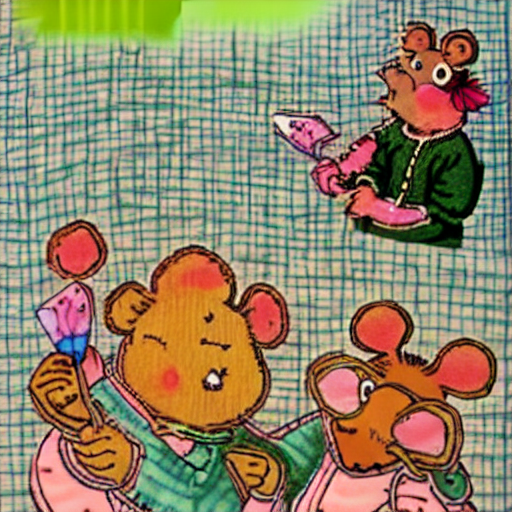} & 
\includegraphics[width=\mywidth,  ,valign=m, keepaspectratio,] {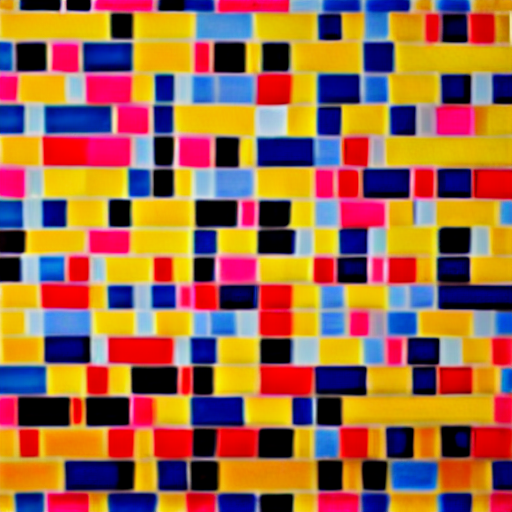} \\

{\rotatebox[origin=t]{90}{\textit{\textbf{\tiny DEADiff}}}} &
\includegraphics[width=\mywidth,  ,valign=m, keepaspectratio,] {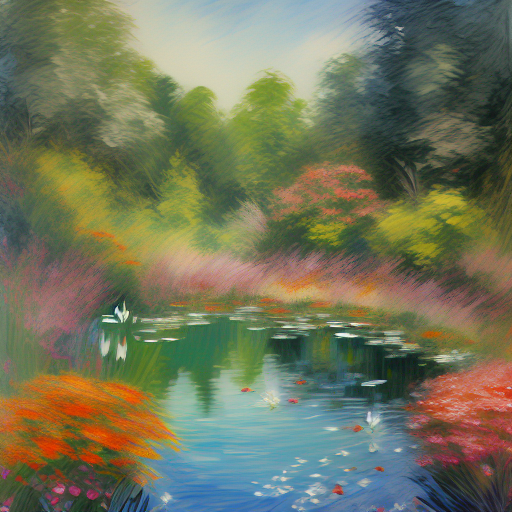} &
\includegraphics[width=\mywidth,  ,valign=m, keepaspectratio,] {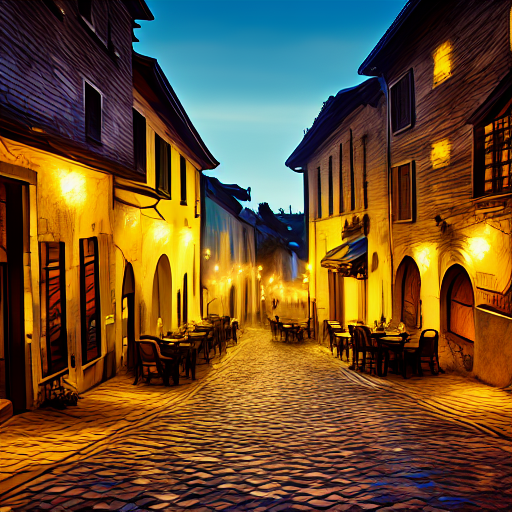} &
\includegraphics[width=\mywidth,  ,valign=m, keepaspectratio,] {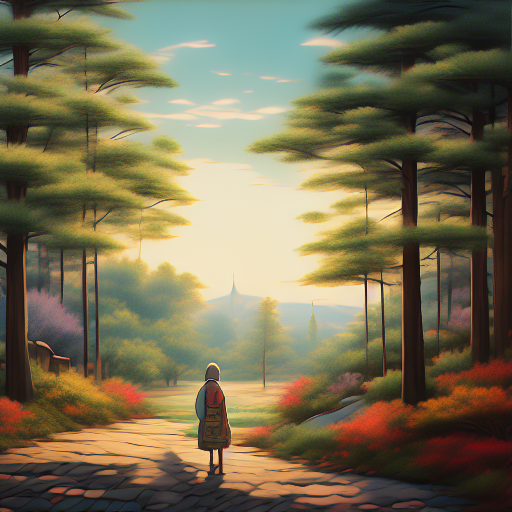} & 
\includegraphics[width=\mywidth,  ,valign=m, keepaspectratio,] {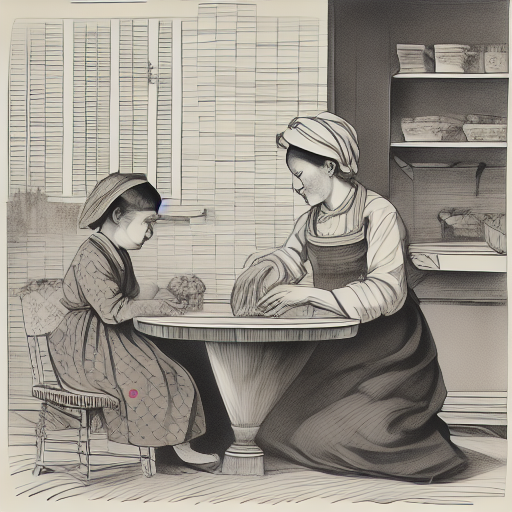} & 
\includegraphics[width=\mywidth,  ,valign=m, keepaspectratio,] {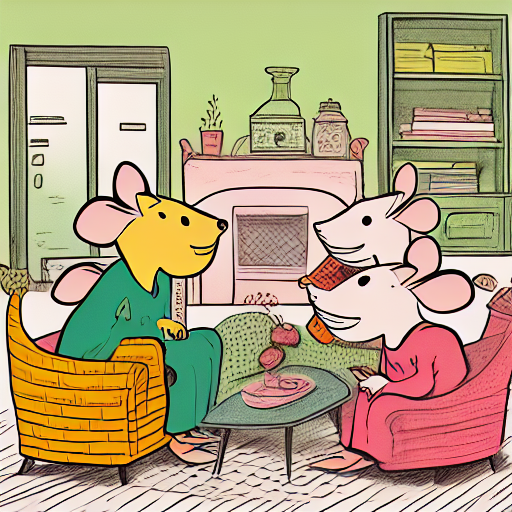} & 
\includegraphics[width=\mywidth,  ,valign=m, keepaspectratio,] {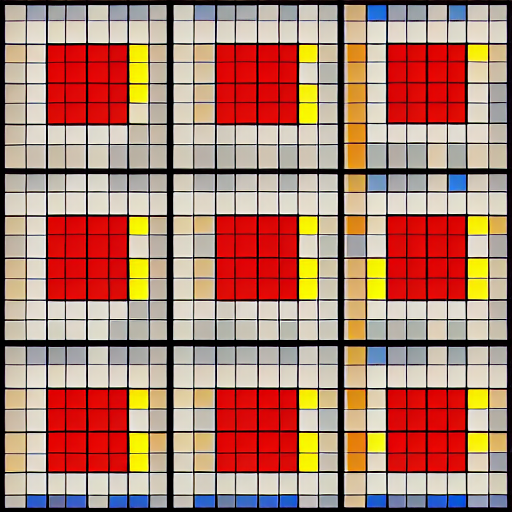} \\

{\rotatebox[origin=t]{90}{\textit{\textbf{\tiny TI}}}} & 
\includegraphics[width=\mywidth,  ,valign=m, keepaspectratio,] {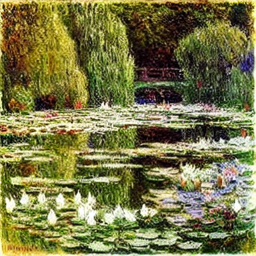} &
\includegraphics[width=\mywidth,  ,valign=m, keepaspectratio,] {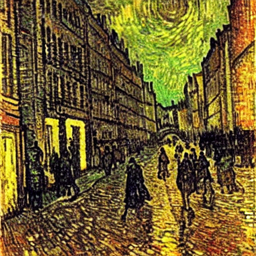} &
\includegraphics[width=\mywidth,  ,valign=m, keepaspectratio,] {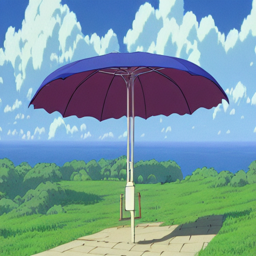} & 
\includegraphics[width=\mywidth,  ,valign=m, keepaspectratio,] {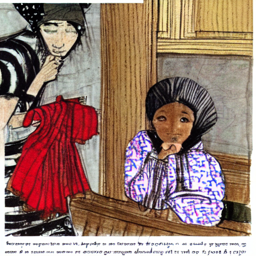} & 
\includegraphics[width=\mywidth,  ,valign=m, keepaspectratio,] {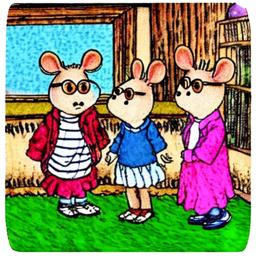} & 
\includegraphics[width=\mywidth,  ,valign=m, keepaspectratio,] {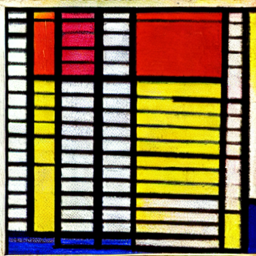} \\

{\rotatebox[origin=t]{90}{\textit{\textbf{\tiny NSYNC}}}} & 
\includegraphics[width=\mywidth,  ,valign=m, keepaspectratio,] {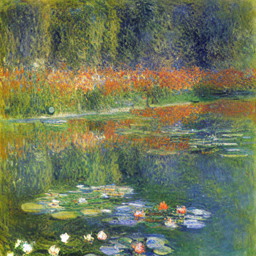} &
\includegraphics[width=\mywidth,  ,valign=m, keepaspectratio,] {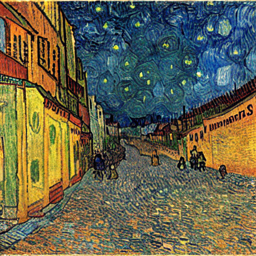} &
\includegraphics[width=\mywidth,  ,valign=m, keepaspectratio,] {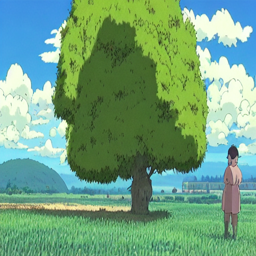} & 
\includegraphics[width=\mywidth,  ,valign=m, keepaspectratio,] {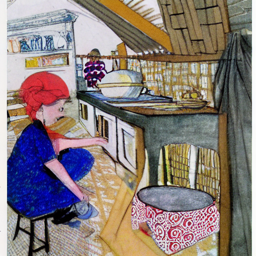} & 
\includegraphics[width=\mywidth,  ,valign=m, keepaspectratio,] {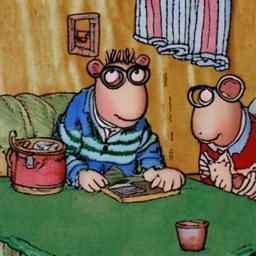} & 
\includegraphics[width=\mywidth,  ,valign=m, keepaspectratio,] {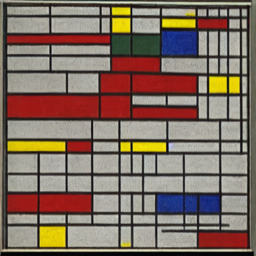} \\
\midrule
{\rotatebox[origin=t]{90}{\textit{\textbf{\tiny GT}}}} & 
\includegraphics[width=\mywidth,  ,valign=m, keepaspectratio,] {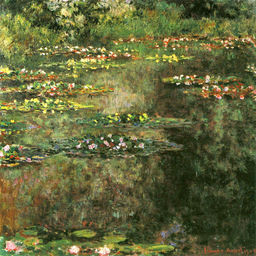} &
\includegraphics[width=\mywidth,  ,valign=m, keepaspectratio,] {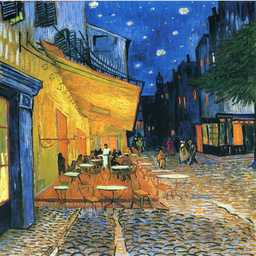} &
\includegraphics[width=\mywidth,  ,valign=m, keepaspectratio,] {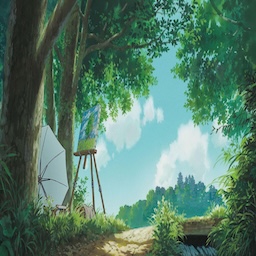} & 
\includegraphics[width=\mywidth,  ,valign=m, keepaspectratio,] {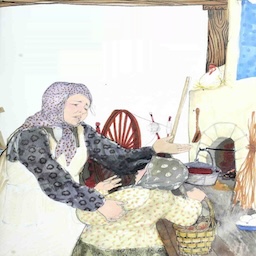} & 
\includegraphics[width=\mywidth,  ,valign=m, keepaspectratio,] {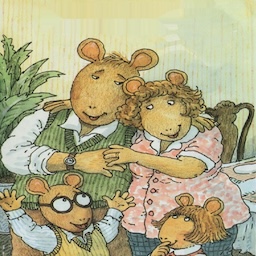} & 
\includegraphics[width=\mywidth,  ,valign=m, keepaspectratio,] {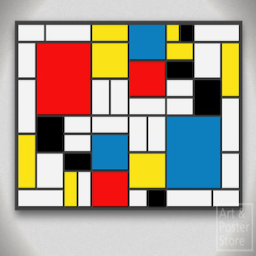} \\

\end{tabular}}
\caption{Comparison of NSYNC with baseline methods using SD v1.x backbones. Ground Truth (GT) is given on the bottom, and textual description to generate the images in given on the top. Note that ground truth (GT) images are for illustration purposes. Zoom in for details.}
\label{fig:comparison}
\end{figure*}
\renewcommand{\arraystretch}{1}

\renewcommand{\arraystretch}{1}
\begin{figure*}
\captionsetup[subfigure]{labelformat=empty}
\centering
\setlength\tabcolsep{1.5pt} 
\resizebox{1.0\textwidth}{!}{
\begin{tabular}{l @{}m{0.20\columnwidth}cc|@{}m{0.20\columnwidth}cc}

{\textit{\textbf{\tiny{DSet}}}} & {~~\textit{\textbf{\tiny{Input}}}} & {\textit{\textbf{\tiny{NSYNC}}}} & {\textit{\textbf{\tiny{GT}}}} & 
{~~\textit{\textbf{\tiny{Input}}}}  & {\textit{\textbf{\tiny{NSYNC}}}} & {\textit{\textbf{\tiny{GT}}}} \\

{\rotatebox[origin=t]{90}{\textit{\textbf{\tiny M}}}} & \tiny{Waves crash against cliffs, silhouettes of two figures stand on the shore.} &
\includegraphics[width=\myw,  ,valign=m, keepaspectratio,] {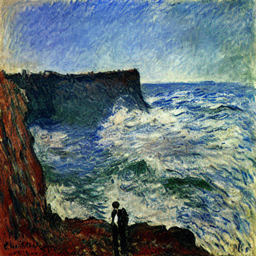} &
\includegraphics[width=\myw,  ,valign=m, keepaspectratio,] {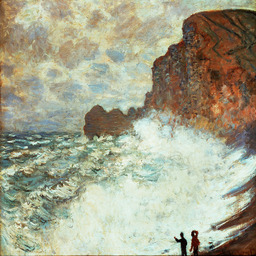} &
\tiny{A serene river flows through a quaint village with charming buildings and a church with twin spires.} & 
\includegraphics[width=\myw,  ,valign=m, keepaspectratio,] {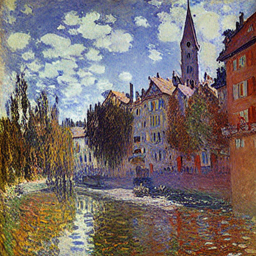} & 
\includegraphics[width=\myw,  ,valign=m, keepaspectratio,] {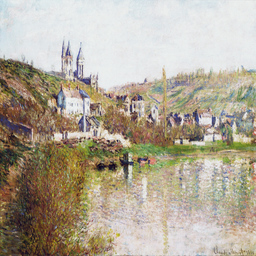}\\

{\rotatebox[origin=t]{90}{\textit{\textbf{\tiny VG}}}}  & \tiny{A vibrant, impressionistic painting of a rural landscape with a stream, houses, and a figure.} &
\includegraphics[width=\myw,  ,valign=m, keepaspectratio,] {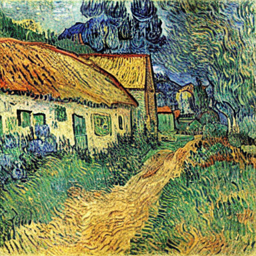} &
\includegraphics[width=\myw,  ,valign=m, keepaspectratio,] {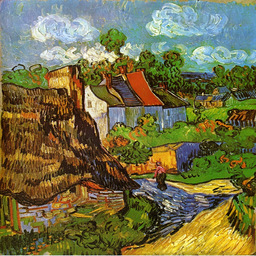} &
\tiny{A moonlit village scene with a cottage, trees, and a sheep.} &
\includegraphics[width=\myw,  ,valign=m, keepaspectratio,] {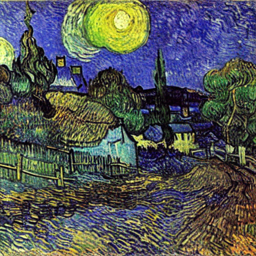}  & 
\includegraphics[width=\myw,  ,valign=m, keepaspectratio,] {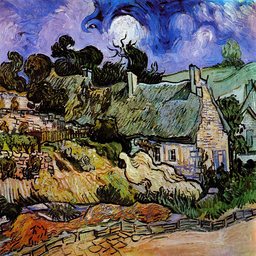}\\

{\rotatebox[origin=t]{90}{\textit{\textbf{\tiny SG}}}} & \tiny{A busy office filled with people working on various designs, using computers and drawing tools.} &
\includegraphics[width=\myw,  ,valign=m, keepaspectratio,] {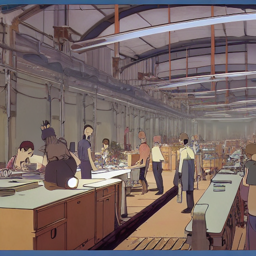} &
\includegraphics[width=\myw,  ,valign=m, keepaspectratio,] {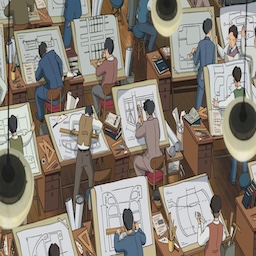} &
\tiny{A colorful, whimsical street scene with buildings adorned with stars and lanterns, a man and two children walking, and a quaint, inviting atmosphere.} & 
\includegraphics[width=\myw,  ,valign=m, keepaspectratio,] {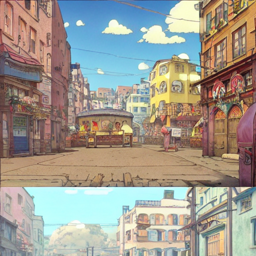}  & 
\includegraphics[width=\myw,  ,valign=m, keepaspectratio,] {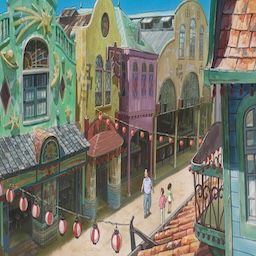}\\

{\rotatebox[origin=t]{90}{\textit{\textbf{\tiny PP}}}} & \tiny{An elderly woman in a red and black floral dress is sewing a white cloth on a wooden table. The room has a patterned rug, a candle, and a window with colorful curtains.} &
\includegraphics[width=\myw,  ,valign=m, keepaspectratio,] {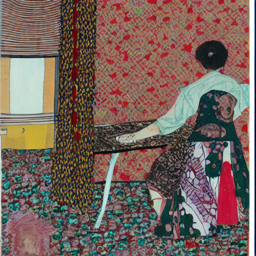} &
\includegraphics[width=\myw,  ,valign=m, keepaspectratio,] {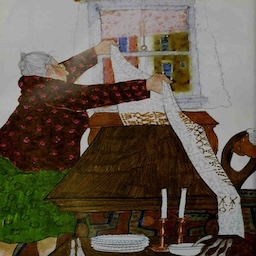} &
\tiny{A woman in a red headscarf sits by a window, reading a book. A small blue bird sits on her lap. The room is cozy, with a cup of tea on the table.} &
\includegraphics[width=\myw,  ,valign=m, keepaspectratio,] {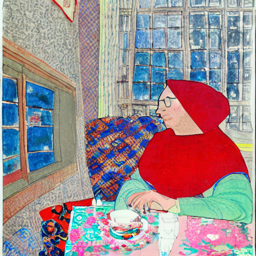} & 
\includegraphics[width=\myw,  ,valign=m, keepaspectratio,] {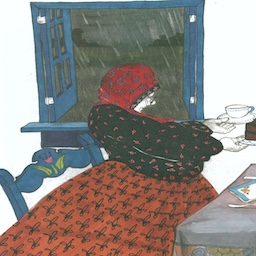} \\

{\rotatebox[origin=t]{90}{\textit{\textbf{\tiny MB}}}} & \tiny{A cartoon depicts a school scene with animals in a classroom. The rabbit is wearing glasses and a yellow sweater, while the horse is in a suit and tie. The background shows students and a teacher.} &
\includegraphics[width=\myw,  ,valign=m, keepaspectratio,] {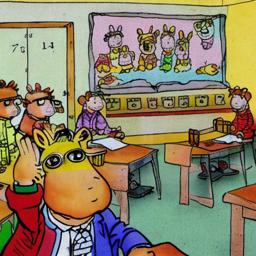} &
\includegraphics[width=\myw,  ,valign=m, keepaspectratio,] {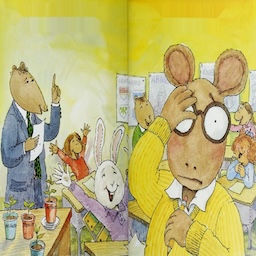} &
\tiny{A woman in glasses and a red cardigan holds a small rabbit. Two children, one in pajamas and the other in a pink dress, stand nearby.} & 
\includegraphics[width=\myw,  ,valign=m, keepaspectratio,] {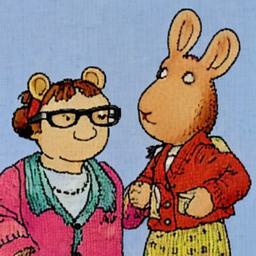}  & 
\includegraphics[width=\myw,  ,valign=m, keepaspectratio,] {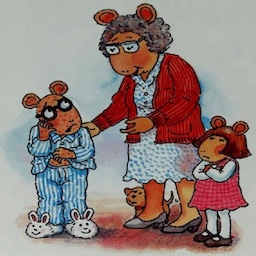}\\

{\rotatebox[origin=t]{90}{\textit{\textbf{\tiny PM}}}} & \tiny{A grid of squares in various colors, including red, yellow, blue, and black.} &
\includegraphics[width=\myw,  ,valign=m, keepaspectratio,] {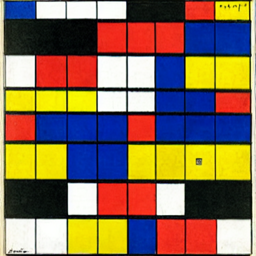} &
\includegraphics[width=\myw,  ,valign=m, keepaspectratio,] {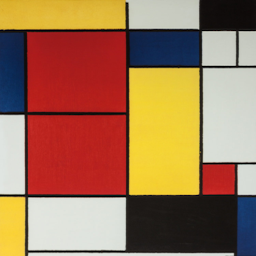} &
\tiny{The image features a grid of squares in red, blue, white, and black.} & 
\includegraphics[width=\myw,  ,valign=m, keepaspectratio,] {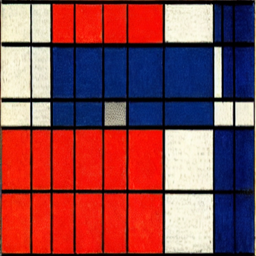}  & 
\includegraphics[width=\myw,  ,valign=m, keepaspectratio,] {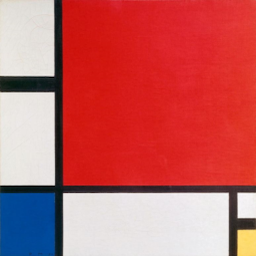}\\

\end{tabular}}
\caption{Visual results of NSYNC on all the datasets. Each row corresponds to one dataset. Ground Truth (GT) is given on the right. Note that ground truth (GT) images are for illustration purposes. Zoom in for details.}
\label{fig:our_results}
\end{figure*}
\renewcommand{\arraystretch}{1}

\subsection{Qualitative Results}

We visually compare the images generated by our approach with those produced by baseline methods in Figure~\ref{fig:comparison} across all datasets.  
Although the generated images are in line with the input descriptions, SD fails to capture the style. 
Moreover, in some cases, it generates natural images instead of stylized images. Similarly, DEADiff also generates natural images that, in most cases, lack any style components. TI significantly improves upon the SD baseline by capturing the style to some extent, especially on illustration and animation datasets. 
However, it lacks specific style elements, such as the brushstrokes characteristic of Monet and Van Gogh or the detailed character illustrations. 
On the other hand, NSYNC results are compliant with the input captions and reflect the style of specific artists, such as Monet's and Cezanne's brushstrokes and Van Gogh's detailed skies.

In Figure~\ref{fig:our_results}, we present further examples to illustrate the capabilities as well as the shortcomings of our model. On painters, the style is successfully transferred; the strokes and color palette of Monet and the abstract shapes of Van Gogh are captured. 
Similarly, on the Studio Ghibli dataset, the style and color palette are completely captured, and the generated images are faithful to the input descriptions.    
On the other hand, in some cases, illustrations fail to comply with the input description, and the model may struggle to generate all the elements. The high CMMD and FID metrics on illustration datasets also support this situation. Nonetheless, the drawing style of the illustrator is evident in the generated images. Again, abstract art results are more or less inline with the input descriptions.

\subsection{Computational Cost}
\label{sec:compute_cost}

We measured the total training time and GPU memory usage for all methods using the same machine equipped with a single NVIDIA A6000 GPU. All models are trained to generate images of the same resolution ($512\times512$); however, they use different default batch sizes. The batch size and corresponding GPU memory usage were as follows: NSYNC and TI 4 / 5376 MB, InST 1 / 11896 MB, and StyleShot 16 / 27800 MB. The training times on the Monet dataset are: NSYNC 1h45m, TI 1h05m, InST 0h45m, and StyleShot 5h15m.

\begin{table*}
\begin{center}
\begin{tabular}{cccc|cccc}
 \toprule 
\multirow{3}{*}{Variants}  &  \multicolumn{3}{c}{Features}     &  \multicolumn{4}{c}{Metrics} \\
 \cmidrule(lr){2-4}  \cmidrule(lr){5-8}
  &  Contrastive Training & Anchor & Gradient Projection & CSD$\uparrow$ & CMMD$\downarrow$ & KID$\downarrow$ & FID$\downarrow$   \\
\midrule 
TI &  \XSolidBrush & \XSolidBrush & \XSolidBrush    & 0.6374 & 0.964 & 0.045 & 163.9      \\
CTM &   \Checkmark & \XSolidBrush  & Mean           & 0.6029 & 0.830 & 0.034 & 153.8    \\
CTMA &   \Checkmark & \Checkmark & Mean             & 0.5943 & 0.820 & 0.024 & 147.3     \\
CTO &   \Checkmark & \XSolidBrush & Orthogonal      & 0.6251 & 0.879 & 0.032 & 153.6   \\
CTOA (NSYNC) & \Checkmark & \Checkmark & Orthogonal & \textbf{0.6484} & \textbf{0.721}  &  \textbf{0.023} & \textbf{139.5}     \\
 \bottomrule 
\end{tabular}
\end{center}
\vspace{-2mm}
\caption{Ablation experiments using different contrastive approaches. All models are on Monet dataset.}
\label{tab:ablations}
\vspace{-4mm}
\end{table*}

\subsection{Ablation Experiments}
\label{sec:ablation}

We explored different approaches to guide the contrastive training. We mainly explored the effect of; 1) using the additional positive \textit{Anchor} image, and 2) using the orthogonal component of the positive and negative gradients. For comparison to the latter case, we simply take the mean of both gradients. We change the prompt by adding a negative style prompting for the negative set (\textbf{not} in the style of $S^*$). Here, our intuition is that when we use a negative prompt for the negative set, its gradient should be in the same direction as the positive one. We refer to this case as using \textit{Mean} gradient projection. Recall that our original approach uses the \textit{Orthogonal} component.
Results of ablation experiments are presented in Table~\ref{tab:ablations} for the Monet dataset. Here \textit{CT} stands for \textit{Contrastive Training}, \textit{M} stands for \textit{Mean} and \textit{O} stands for \textit{Orthogonal} gradient projection, and \textit{A} stands for using \textit{Anchor} images. Note that NSYNC corresponds to CTOA.

Using an additional \textit{Anchor} image improves the baseline for both gradient update mechanisms. It especially boosts the performance on the orthogonal gradient projection case due to a more accurate calculation of the positive gradient direction. Even using a simple gradient \textit{Mean} approach along with negative prompting performs significantly better than TI~\cite{ti_model} model, showing the importance of having a negative set.

\section{Conclusion}
\label{sec:conc}

We introduced a new approach to utilize synthetic images in image generation tasks. Specifically, we propose to generate synthetic images to resemble the negative set of our target images. The proposed  contrastive representation learning framework aims to better guide the image generation process. Negative sets are curated by injecting negative prompts into the image captions. Our method NSYNC, finetunes the baseline adapter model~\cite{ti_model} by measuring the gradient directions coming from positive and negative sets and updates the weights of the model accordingly. Experiments on painting and illustration datasets showed the effectiveness of our model both quantitatively and visually. In the future, we plan to show the effectiveness of negative set generation in other vision tasks.
\section{Future Work}
\label{sec:future}

\noindent{\textbf{Diversity of Synthetic Negatives}}
The current approach, i.e., using a single prompt for negative set curation, has limitations. However, a thorough investigation and quantification of this phenomenon would require a dedicated, in-depth analysis, which is beyond the scope of our current work. We acknowledge this important direction and list it as part of our future work.

\noindent{\textbf{Other Style Forms}}
In this work, we experimented with four different art domains: paintings, animations, illustrations, and abstract art. Our work could be extended to other forms of art, such as photographic aesthetics. We leave the support for other fine-grained sets of styles as future work.

\section*{Acknowledgments}
The numerical calculations reported in this paper were fully performed at TUBITAK ULAKBIM, High Performance and Grid Computing Center (TRUBA resources). This work is supported by the Turkish Science Academy (BAGEP).

\bibliographystyle{IEEEtran}
\bibliography{sample-base}

\end{document}